# Optimal Event Monitoring through Internet Mashup over Multivariate Time Series

Chun-Kit Ngan, George Mason University, USA

Alexander Brodsky, George Mason University, USA

## ABSTRACT

We propose a Web-Mashup Application Service Framework for Multivariate Time Series Analytics (MTSA) that supports the services of model definitions, querying, parameter learning, model evaluations, data monitoring, decision recommendations, and web portals. This framework maintains the advantage of combining the strengths of both the domain-knowledge-based and the formal-learning-based approaches and is designed for a more general class of problems over multivariate time series. More specifically, we identify a general-hybrid-based model, MTSA – Parameter Estimation, to solve this class of problems in which the objective function is maximized or minimized from the optimal decision parameters regardless of particular time points. This model also allows domain experts to include multiple types of constraints, e.g., global constraints and monitoring constraints. We further extend the MTSA data model and query language to support this class of problems for the services of learning, monitoring, and recommendation. At the end, we conduct an experimental case study for a university campus microgrid as a practical example to demonstrate our proposed framework, models, and language.

## INTRODUCTION

Observing behaviors, trends, and patterns on multivariate time series (Bisgaard & Kulahci, 2011; Chatfield, 2001) has been broadly used in various application domains, such as financial markets, medical treatments, economic studies, and electric power management. Domain experts utilize multiple time series to detect events and make better decisions. For example, financial analysts predict different states of the stock market, e.g., bull or bear, more accurately based upon monitoring daily stock prices, weekly interest rates, and monthly price indices. Physicians monitor patients' health conditions by measuring their diastolic and systolic blood pressures, as well as their electrocardiogram tracings over time. Sociologists uncover hidden social problems within a community more profoundly through studying a variety of economic, medical, and social indicators, e.g., annual birth rates, mortality rates, accident rates, and various crime rates. The goal of examining those characteristics over multivariate time series on events is to support decision makers, e.g., financial analysts, physicians, and sociologists, to better understand a problem in different perspectives within a particular domain and to offer better actionable recommendations.

To support such an event-based decision-making and determination over multivariate time series, in this paper, we propose a Web-Mashup Application Service Framework for Multivariate

Time Series Analytics (MTSA). This framework is an integrated tool to support the MTSA service development, including model definitions, querying, parameter learning, data monitoring, decision recommendations, and model evaluations. Domain experts could use the framework to develop and implement their web-based decision-making applications on the Internet. Using a Web Mashup function offered by the Web 2.0 technology (Vancea & Others, 2008; Gurram & Others, 2008; Murugesan, 2007;Bradley, 2008; Alonso & Others, 2004; Altinel & Others, 2007; Ennals & Others, 2007;Thor & Others, 2007) on our framework, domain experts could collect and unify global information and data from different channels and media, such as web sites, data sources, organizational information, etc., to generate a concentric view of collected time series data from which the learning service determines optimal decision parameters. Using optimal decision parameters, domain experts can employ the monitoring service to detect events and the recommendation service to suggest actions.

Presently, there are two key approaches that domain users utilize to identify and detect interesting events over multivariate time series. These approaches are domain-knowledge-based and formal-learning-based. The former approach completely relies on domain experts' knowledge. Based on their knowledge and experience, domain experts determine monitoring conditions that detect events of interest and trigger an appropriate action. More specifically, domain experts, e.g., financial analysts, have identified several deterministic time series, such as the S&P 500 percentage decline (SPD) and the Consumer Confidence Index drop (CCD), from which they develop parametric monitoring templates, e.g., SPD < -20%, CCD < -30 (Stack, 2009), etc., according to their expertise. Once the incoming time series, i.e., SPD and CCD, satisfy the given templates at a particular time point, the financial analysts decide that the bear market bottom is coming, which is the best buy opportunity to purchase the stock to earn the maximal earning.

Consider another real-world case study of the timely event detection of certain conditions in the electric power microgrid at George Mason University (GMU), where its energy planners would like to regularly detect when the electric power demand (*electricPowerDemand*) exceeds the pre-determined peak demand bound (*peakDemandBound*). The reason is that the occurrence of this event leads to a significant portion of the GMU electric bill based upon its contractual terms even though the event, *electricPowerDemand* > *peakDemandBound*, occurs only within a short period of time, e.g., one minute. Thus such an identification and detection can aid in the task of decision-making and the determination of action plans. To make better decisions and determinations, the energy planners have identified a set of time series that can be used to detect the event and perform an action, e.g., to execute the electric load shedding to shut down some electric account units on the GMU campus according to a prioritization scheme from the energy manager. The multiple time series include the input electric power demand per hourly time interval, the given peak demand bound per monthly pay period, etc. If these time series satisfy a pre-defined, parameterized condition, e.g., *electricPowerDemand* > *peakDemandBound*, where the given *peakDemandBound* is 17200 kWh for all the hourly time intervals within the same monthly pay period, e.g., July, 2012, it signals the energy planners to execute the electric load shedding in the microgrid on the campus. Often these parameters, e.g., the predetermined peak demand bound, may reflect some realities since they are set by domain experts, e.g., the energy planners, based on their past experiences, observations, intuitions, and domain knowledge. However, these given thresholds, e.g., the peak demand bound, are not always accurate. In addition, the parameters are static, but the problem that we deal with is often dynamic in nature, so the parameters definitely are not the optimal values for achieving the monitoring purpose at different periods of time, e.g., hourly, daily, monthly, quarterly, and yearly, to minimize the electricity expenses of the bill. Thus this domain-knowledge-based approach lacks a formal mathematical foundation that dynamically learns optimal decision parameters to determine an event.

The latter approach utilizes a formal learning methodology, such as a non-linear logistic regression model (Bierens, 2008; Cook & Others, 2000; Dougherty, 2007; Hansen, 2010; Heij & Others, 2004). This regression model is used to predict the occurrence of an event (0 or 1), e.g., when to shed load or unshed load, by learning parametric coefficients of the logistic distribution function of explanatory variables, i.e., the electric power demand and the peak demand bound. More specifically, this non-linear logistic regression model focuses on modeling the data relationship between explanatory variables and response variables. The truth is that not all the response variables are numeric and continuous. In many real-world cases, the responses may only take one of two possible answers, e.g., shed load or unshed load, buy or sell stocks, success or failure, etc. Each outcome of the responses is assigned to a value 1 if the probability of the event happening is above 0.5 and 0 otherwise. To learn the parametric coefficients of the logistic distribution function of explanatory variables to determine the outcome of the binary responses, we can apply the nonlinear logistic regression model and the Maximum Likelihood Estimation (MLE) (Myung, 2003) over historical and projected data. However, the main challenge of using formal learning approaches is that they do not always produce satisfactory results, as they do not consider incorporating domain knowledge, including monitoring constraints, e.g., $electricPowerDemand > peakDemandBound$ , and global constraints, e.g., utility contractual terms, into their formal learning aproaches. Lacking domain experts' knowledge on parameter learning will result in an inaccurate decision-making. For instance, the energy planners might execute the electric load shedding at an improper moment of time, particularly during the business office hours between 9:00 a.m. and 6:00 p.m. from Monday to Friday.

Some existing mathematical models, e.g., the Durland and McCurdy duration-dependent Markov-switching (DDMS) models, such as DDMS-ARCH and DDMS-DD (Maheu and McCurdy, 2000), do integrate domain knowledge, e.g., duration dependence, into their forecasting criteria. Both models, DDMS-ARCH and DDMS-DD, are extended from the Markov-switching model (Bickel, et al., 1998) that is incorporated with duration dependence to affect a transition probability that is parameterized using the logistic distribution function. The transition probability is the probability of being in a particular state at a specific point in time. The value and the trend of this probability over time demonstrates the current state of an event. However, all of these models only consider a single element, i.e., duration, to integrate into the model to determine a state of an event. This approach is not flexible and complete as there are many other external, unknown factors that may affect the state of an event in the currenct environment. In addition, those models also involve parameters that need to be learnt by formal mathematical computations. Without wide-ranging domain experts' knoweledge, those formal learning methods become computationally intensive and time consuming. The whole model building is an iterative and interactive process, including model formulation, parameter estimation, and model evaluation. Despite enormous improvements in computer software in recent years, fitting such nonlinear quantitative decision model (Evans, 2010) is not a trival task, especially if the parameter learning process involves multiple explanatory variables, i.e., high dimensionality. Moreover, working with high-dimensional data creates difficult challenges, a phenomenon known as the "curse of dimensionality" (Bellman, 1957 and 1961). Specifically, the amount of observations required in order to obtain good estimates increases exponentially with the increase of dimensionality. In addition, many learning algorithms do not scale well on high dimensional data due to the high computational cost. The parameter computations by formal-learning-based approaches, e.g., logistic regression model, are complicated and costly, and they lack the consideration of integrating various experts' domain knowledge into the learning process – a step that could potentially reduce the dimensionality. Clearly, both approaches, domain-knowledge-based and formal-learning-based, do not take advantage of each other to learn optimal decision parameters, which are then used to monitor the events and to take appropriate actions.

To mitigate the shortcomings of the existing approaches, we have proposed a mathematical hybrid-based model, Expert Query Parametric Estimation (EQPE), and an SQL-based language (Ngan, Brodsky & Lin, 2010), which combine the strengths of both domain-knowledge-based and formal-learning-based approaches. More specifically, we take a monitoring template of conditions in a specific form, that is, conjunctions of inequality constraints, identified by domain experts. This template consists of inequalities of values in time sequences and then is parameterized. The goal is to find parameters that maximize or minimize an objective function in which the function is depended on optimal time points of a time utility function from which the parameters are learned. Because of these characteristics, however, the EQPE model is only able to solve a specific class of problems that (1) their decision parameters of an objective function are learned from optimal time points of a time utility function, (2) the monitoring template has to be in the considered form, i.e., conjunctions of inequality constraints, only, and (3) the constraints being used are solely for monitoring purposes.

To address the above weaknesses, the proposed web-mashup application service framework for MTSA also maintains the advantage of combining the strengths of both the domain-knowledge-based and the formal-learning-based approaches, but it is designed for a more general class of problems over multivariate time series. This service framework supports quick implementations of services towards decision recommendations on events. More specifically, the *MTSA Model Definition Service* takes multiple templates of conditions, for example, the monitoring template to determine the occurrence of an event identified by domain experts, the general template for a contractual term of an electric bill required by power companies, etc. Such templates consist of inequalities of values in time sequences, and then the *Learning Service* "parameterizes" it, e.g., *electricPowerDemand* > *peakDemandBound*. The goal of the learning service is to learn parameters that optimize the objective function, e.g., minimizing the cost of the GMU electric bill. The *Monitoring and Recommendation Service* continuously monitors the data streams that satisfy the parameterized conditions of the monitoring template, in which the parameters have been instantiated by the learning service.

To support such services for a general class of problems, we further extend the proposed relational database model and SQL with high-level MTSA constructs. This further extension can support parameter learning, data monitoring, and decision recommendation over multivariate time series for this class of problems. To this end, we identify a general-hybrid-based model, Multivariate Time Series Analytics – Parameter Estimation (MTSA-PE). This model is a combination of both domain-knowledge-based and formal-learning-based approaches with possibly incorporating any global constraints, e.g., the contractual terms of the GMU electric bill, which are applied to an entire problem, and monitoring constraints, e.g., *electricPowerDemand* > *peakDemandBound*, which are used to detect the occurrence of an event. Both types of inequality constraints, global and monitoring, are allowed in any possible combinations and forms. Using the MTSA-PE model, domain experts can learn decision parameters that satisfy all the given constraints and that optimize the objective function, which is independent of a particular time point.

To demonstrate our MTSA-PE model, we conduct an experimental case study on the Fairfax campus microgrid at GMU. We utilize the MTSA-PE model to illustrate the GMU problem and the further extended MTSA-query constructs to express the model. After the MTSA-query constructs are initiated to learn the optimal peak demand bound over historical and projected electric power demands, the occurrence of the event can be monitored and determined through the parametric monitoring constraints, e.g., *electricPowerDemand* > *peakDemandBound*. Once the event is detected, the electric load shedding can be executed.

The rest of the paper is organized as follows: using the GMU Fairfax campus microgrid as an example, we describe its electric bill problem in the second section. In the third section, we provide an overview on the web-mashup application service framework for multivariate time series analytics and describe its supports of quick service implementations towards recommendations on events over multivariate time series. In the fourth section, we use the GMU electric bill as an example to describe the further extended MTSA data model and query language that is used for the MTSA service implementations of the general class of problems. We also define the Multivariate Time Series Analytics - Parameter Estimation (MTSA-PE) model for the MTSA-query semantics and use the GMU electric bill problem to illustrate the learning, monitoring, and recommendation services on this model in the fifth section. In the sixth section, we present the architecture for the parameter learning process. In the seventh section, we conduct and describe the experimental case study on that GMU problem. In the eighth section, we conclude and briefly outline the future work.

## PROBLEM DESCRIPTION OF A REAL CASE STUDY

Consider the real case study at George Mason University (GMU), where the electric power demand across the expanding Fairfax and other campuses is expected to increase. The increase in power consumption results in a higher electricity cost, which is composed of the two main components: (1) a total kilowatt-hour (kWh) charge, i.e., the charge for the total electricity consumption, and (2) an Electricity Supply (ES) service charge, i.e., the charge for the peak demand usage in any 30-minute interval over the past 12 months. The total kWh charge is priced particularly higher during the business office hours between 09:00 a.m. and 06:00 p.m. from Monday to Friday. This monthly ES service charge (*monthlyEServiceCharge*) is a proxy for the cost of capital investment for power generation capacity, since the power company, Virginia Electric and Power Company, needs to build generation, transmission, and distribution facilities that are capable of supporting the peak demand, even though the average power demand could be considerably lower. This ES service charge amounts to approximately 30% of the electric bill in each monthly pay period (*payPeriod*) and is determined based upon the electricity supply demand (*payPeriodSupplyDemand*). This electricity supply demand is decided on the highest of either (*C1*) or (*C2*):

*C1*: The highest average kilowatt measured in any 30-minute interval of the current billing month during the on-peak hours of either:

- Between 10 a.m. and 10 p.m. from Monday to Friday for the billing months of June through September or
- Between 7 a.m. and 10 p.m. from Monday to Friday for all other billing months.

*C2*: 90% of the highest kilowatt of demand at the same location as determined under (*C1*) above during the billing months of June through September of the preceding eleven billing months.

Thus it is possible that a high peak demand usage just for one minute of electricity consumption over the past year would result in a very significant increase in the charge of the electric bill. Therefore, controlling the peak demand is crucial for controlling the electricity cost. However, an important question is how the commercial and industry customers, such as GMU, should respond to those contractual terms of the electric bill mentioned above.

Our key idea is to learn an optimal peak demand bound over historical and projected electric power demands for each future pay period. This optimal bound is then used to monitor the prospective demand usage in any time interval of that future pay period. Once the demand usage exceeds the bound, some electric loads are shed to shut down some electric account units so that the ES service charge can be controlled. However, to determine an optimal peak demand bound is challenging, as if the bound is set too high, although power services are not interrupted, customers will be charged a significant electricity expense. If the bound is set too low, a low electricity charge is billed, but more power interruptions to customers occur. In order to make an optimal balance of this trade-off, we propose a web-mashup application service framework for multivariate time series analytics that are designed for domain experts to solve this dilemmatic situation.

## A WEB-MASHUP APPLICATION SERVICE FRAMEWORK FOR MULTIVARIATE TIME SERIES ANALYTICS

Figure 1 shows a range of common web services which is desirable to be offered over the Internet to address the problems with the one that is described in the second section. The *MTSA Model Definition Service* allows domain experts to define different types of parametric model templates, which are identified by the experts and/or are required by the contracts. For example, in the GMU case study, the energy planners would like to detect when the electric power demand at an hourly time interval exceeds a predetermined peak demand bound. The designed model template consists of time series, including the electric power demand at an hourly time interval, the peak demand bound for each time interval at a monthly pay period, etc. Some of the time series, e.g., the electric power demand, are the input time series, and some of them, e.g., the peak demand bound, are the time series that are instantiated from the historical and projected electric power demands. These time series are associated with their respective inequality constraints, for example, *electricPowerDemand* > *peakDemandBound*.

Given such a parametric model template in a given domain, the *Monitoring and Recommendation Service* continuously screens the incoming data stream for time series that satisfy all the given constraints. These constraints specify when the event of interest, e.g., *electricPowerDemand* > *peakDemandBound*, has occurred, and then the service recommends an action, e.g., to execute an electric load shedding to shut down some electric account units based upon the prioritization scheme from the energy manager to prevent the new incoming electric power demand from exceeding the peak demand bound. Note that in the traditional approach, the decision parameters, such as the peak demand bound, is specified by domain experts, for instance, the energy planners. However, using such a hard-set parameter cannot capture the dynamics of the rapidly changing electric power consumptions at different periods of time, e.g., hourly, daily, monthly, quarterly, and yearly.

To address the above deficiency, the *Parameter Learning Service* parameterizes a given model template, e.g., *electricPowerDemand* > *peakDemandBound*, from domain experts, and supports learning of decision parameters over historical and projected time series. More specifically, decision parameters, e.g., *peakDemandBound*, on the template are instantiated from the input historical and projected time series, e.g., *electricPowerDemand*, such that the learned parameters not only satisfy the template but also optimize the objective function, for instance, minimizing the ES service charge in a monthly pay period. Then the accuracy of decision parameters can be ensured through the *Model Accuracy and Quality Evaluation Service*, which validates the prediction, e.g., *electricPowerDemand* > *peakDemandBound*, with the observed real data, and updates the model template if it is necessary.

The *Querying Service* allows domain experts to express their complex information services over multivariate time series discussed above in a high-level abstraction. More precisely, using the *Querying Service*, domain experts can code the MTSA SQL-like language to develop and implement any MTSA services, for example, the MTSA-SQL constructs of parameter learning, data monitoring, and decision recommendation services.

The *Web Portal Service* enables domain experts to develop a point-of-access service as a main entrance (1) to integrate all their implemented MTSA services together, (2) to provide a consistent style and format among all those services, and (3) to centralize users' access control and procedures on those services.

*Figure 1. Web Services for Multivariate Time Series Analytics over the Internet*

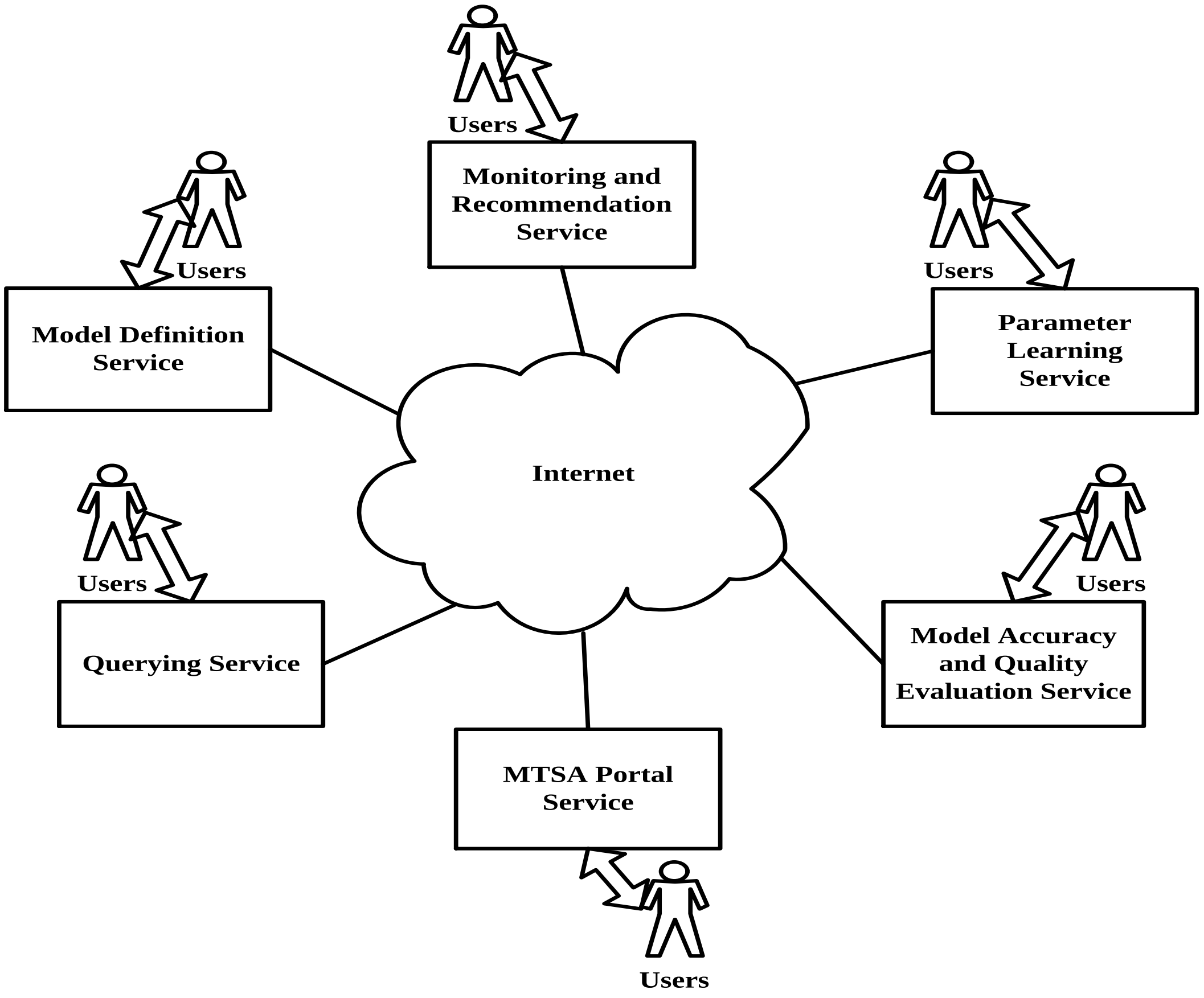

Due to the increasing demand of the MTSA services, we propose a web-mashup application service framework. The web-mashup application service framework is a development tool that provides a medium to domain experts and supports their quick implementations of the services, which are described above. This MTSA service framework is illustrated in Figure 2. It consists of five main components: Data Source Collector (DSC), Data Mashup Integrator (DMI), MTSA Data Model Definition and Query Language Interface (DMD-QLI), MTSA Compiler, and Web Application Designer (WAD).

Integrated with the mashup technology of Web 2.0, the DSC allows domain experts to directly interact with external data services and collect multivariate time series data from heterogeneous sources, including web data, XML documents, enterprise databases, excel/CSV files, WSDL-based web services information, and RSS feeds, around the globe. After multivariate time series are collected by the DSC, domain experts can operate the DMI that is a data integration processing unit to provide a concentric view and maintain a consistency of the collected data, which are then archived in the local databases of multivariate time series.

The DMD-QLI enables domain experts to use the further extended relational database models with the time series and events and SQL with the high-level MTSA constructs. These constructs include MTSA parametric model templates, querying, decision parameter learning, data monitoring and decision recommendations, as well as model evaluations. Using the DMD-QLI, domain experts utilize the MTSA query language (1) to create the further extended relational database models for multivariate time series data, which are used in parametric model templates, (2) to create and initiate learning events, e.g., learning the peak demand bound, which are transformed into the IBM OPL constructs (Hentenryck, 1999; The IBM Corporation, 2012; ILOG S.A. & ILOG, Inc, 2007) by the MTSA compiler, which sends the constructs to the external optimization solver, i.e., the IBM ILOG CPLEX Optimizer, from which the optimal decision parameters on the parametric model templates are learned over historical and projected multivariate time series, (3) to develop and implement the data monitoring and decision recommendation services by the traditional SQLs with the "**MONITOR**" keyword and the learned decision parameters to monitor the events, and (4) to develop a MTSA construct to evaluate model accuracy and quality. Note that those extended relational database models, learned decision parameters, and parametric model templates all are stored in the local database repositories for the future use.

The WAD provides a user-friendly designer that offers domain experts a Java IDE environment and a JSF/JSP work space in which they can develop and implement a web portal and its associated web pages, which directly interact with all the MTSA services developed from the DMD-QLI. Domain experts can also use the web portal to centralize, administrate, and access all the developed MTSA services.

*Figure 2. A Web-Mashup Application Service Framework for Multivariate Time Series Analytics*

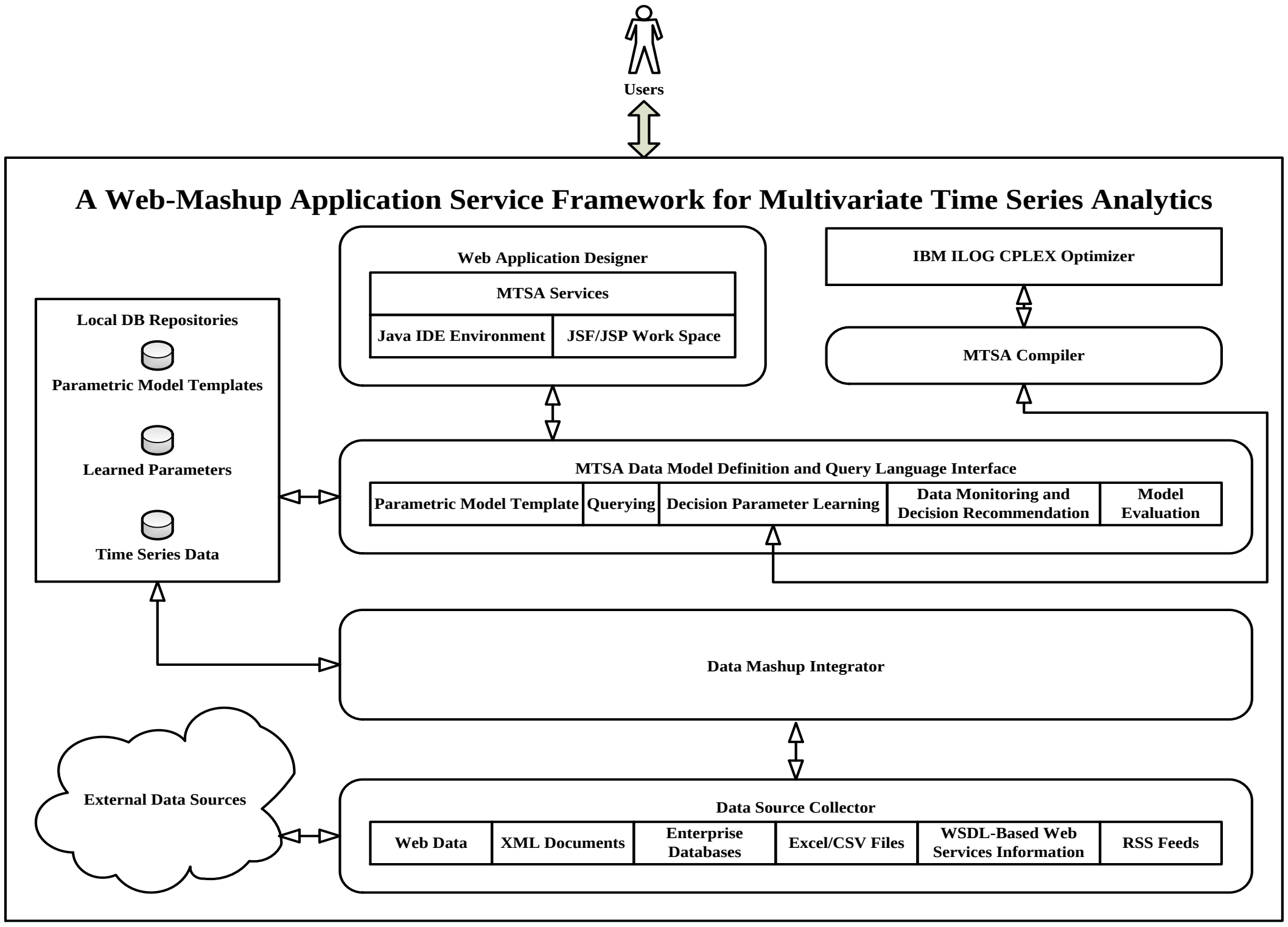

# MTSA DATA MODEL AND QUERY LANGUAGE BY EXAMPLE

## Data Model

Before those MTSA services are developed from the DMD-QLI, a time-series data model for those services is needed to be created. The time-series (TS) data model is a SQL-based extension of the relational database model with the specialized schemas:

- A time-series (TS) schema is of the form TSname(Tname:Ttype, Vname:Vtype). Ttype and Vtype are data types, where Ttype can be any system date/time format or integer-based time interval to show the time sequential data, and Vtype is either Real or Integer. TSname, Tname, and Vname are the attribute names chosen by users.
- A time-event (TE) schema is of the form TEname(Tname:Ttype, Ename:Binary), where Binary is the binary type corresponding to the domain {0,1}, and TEname and Ename are attribute names chosen by users.
- A decision-parameter (DP) schema is of the form DParameter(Tname: Ttype, Pname: Ptype, Vname:Vtype), where Ptype can be any system date/time format or integer-based period interval that is corresponding to a set of time intervals.
- A Time Series database schema is a set of relational schemas which may include TS, TE, and DP schemas.

A TS tuple over a schema TSname(Tname:Ttype, Vname:Vtype) is a relational tuple over that schema, i.e., a mapping m: {Tname, Vname} → Dom(Ttype) x Dom(Vtype), such that m(Tname) ∈ Dom(Ttype) and m(Vname) ∈ Dom(Vtype).

A TE tuple over a similar schema TEname(Tname:Ttype, Ename:Binary) is a mapping m: {Tname, Ename} → Dom(Ttype) x Dom(Binary), such that m(Tname) ∈ Dom(Ttype) and m(Ename) ∈ Dom(Binary).

A DP tuple over a schema DParameter(Tname: Ttype, Pname: Ptype, Vname:Vtype) is a mapping m: {Tname, Pname, Vname} → Dom(Ttype) x Dom(Ptype) x Dom(Vtype), such that m(Tname) ∈ Dom(Ttype), m(Pname) ∈ Dom(Ptype), and m(Vname) ∈ Dom(Vtype).

Let us consider our GMU example. Using the further extended MTSA data model, the energy planners can create the time-series tables as the inputs and stores them with the data in the database. For example, $ElectricPowerDemand(time, value)$ (Box 1) is the input time-series table, and $PeakDemandBound(time, period, value)$ (Box 2) is the parameter table. Both tables are created as follows. We will show the time-event views in the next sub-section.

*Box 1.*

```
CREATE TABLE ElectricPowerDemand (
      time HOURLY_INTERVAL,
      value REAL);
```

*Box 2.*

```
CREATE TABLE PeakDemandBound (
      time HOURLY_INTERVAL,
      period MONTHLY_INTERVAL,
      value REAL,
      UNIQUE_MAP(time, period));
```

`HOURLY_INTERVAL, DAILY_INTERVAL, MONTHLY_INTERVAL, QUARTERLY_INTERVAL, and YEARLY_INTERVAL` are the new integer-based data types to show the sequence of the data. `UNIQUE_MAP()` is the new function that ensures each hourly interval is mapped to one monthly interval, for example. Note that we use the negative and zero integers, e.g., `time` ≤ 0, `period` ≤ 0, etc., indicate the historical time series, and the positive integers, e.g., `time` ≥ 1, `period` ≥ 1 denote the projected time series.

## Monitoring and Recommendation Service

Using the monitoring and recommendation service, the energy planners can determine when they should execute the electric load shedding. In our GMU example, one of the input time series tables is $ElectricPowerDemand(time, value)$ (Box 1), which is created above to store the new incoming electric power demand for monitoring. The monitoring and recommendation service can be expressed by a monitoring-event view and executed by the **MONITOR** command (Box 3 and Box 4).

*Box 3.*

```
CREATE VIEW ElectricLoadShedding AS (
      SELECT EPD.time, (CASE WHEN EPD.value > PDB.value
                  THEN '1' ELSE '0' END) AS Indicator
      FROM  ElectricPowerDemand EPD, PeakDemandBound PDB
```

```
      WHERE EPD.time = PDB.time);
```

*Box 4.*

```
CREATE VIEW ELS_Monitoring_Recommendation AS (
      SELECT ELS.time, (CASE WHEN ELS.Indicator = '1'
                  THEN 'The Electric Power Demand Greater Than The Peak
                  Demand Bound. The Electric Load Shedding Is
            Recommended.' END) AS Action
      FROM ElectricLoadShedding ELS);

MONITOR ELS_Monitoring_Recommendation;
```

`PeakDemandBound` stores the given decision parameter, e.g., 17200 kWh for all the hourly time intervals within the same monthly pay period, e.g., July, 2012. If the monitoring constraint in the "CASE WHEN" clause of the `ElectricLoadShedding` view (Box 3) is satisfied at the current time interval `time`, the value of the attribute "Indicator" indicates "1". The service then recommends the energy planners to execute the electric load shedding since the electric power demand is greater than the peak demand bound (Box 4).

**Parameter Learning Service**

As we discussed, domain experts' suggested parameters are not accurate enough to monitor the dynamics of the rapidly changing electric power consumptions at different periods of time, e.g., hourly, daily, monthly, quarterly, and yearly; thus, the parameter learning service should be adopted to learn the optimal decision parameters, and this service can be expressed as follows:

STEP 1: Store the input Time Series tables, e.g., ElectricPowerDemand, PayPeriod, WeekDay, Hour, Month, etc., in the database.

STEP 2: Create the parameter tables, e.g., PeakDemandBound, PayPeriodSupplyDemand, etc., to store the optimal decision parameters.

STEP 3: Create a Time Series view for the monthly ES service charge for each pay period (Box 5). We assume that the future pay periods are two years, i.e., 24 pay periods as there are 24 months.

*Box 5.*

```
CREATE VIEW MonthlyEServiceCharge AS (
      SELECT PPSD.time, PPSD.period, 8.124 * PPSD.value AS Charge
      FROM PayPeriodSupplyDemand PPSD);
```

$8.124 is the generation demand charge per kilowatt according to the contract of the GMU electric bill.

STEP 4: Create the global constraints, e.g., the condition *C1* of the contractual terms of the GMU electric bill (Box 6), which we described in the second section.

*Box 6.*

```
CREATE VIEW CurrentBillingMonth AS (
      SELECT PayPeriod.time, PayPeriod.period,
            PayPeriodSupplyDemand.value AS payPeriodSupplyDemand,
            KW.value AS kw,
            (CASE WHEN (WeekDay.d >= 1 AND WeekDay.d <= 5)
                  AND ((Hour.h >= 10 AND Hour.h <= 22
                  AND Month.m >= 6 AND Month.m <= 9)
                  OR ((Hour.h >= 7 AND Hour.h <= 22)
                  AND (Month.m <= 5 OR Month.m >= 10)))
                  AND PayPeriod.time = WeekDay.time
                  AND WeekDay.time = Hour.time
                  AND Hour.time = Month.time
                  AND Month.time = PayPeriodSupplyDemand.time
                  AND PayPeriodSupplyDemand.time = KW.time
            THEN '1' ELSE '0' END) AS Indicator
      FROM PayPeriod, WeekDay, Hour, Month, PayPeriodSupplyDemand, KW);
```

STEP 5: Create the monitoring constraints (Box 7), e.g., electricPowerDemand > peakDemandBound.

*Box 7.*

```
CREATE VIEW ElectricPowerPeakDemandBound AS (
      SELECT PayPeriod.time, PayPeriod.period,
            PeakDemandBound.value AS peakDemandBound, KW.value AS kw,
            (CASE WHEN ElectricPowerDemand.value > PeakDemandBound.value
                  AND PayPeriod.time >= 1
                  AND PayPeriod.time = ElectricPowerDemand.time
                  AND ElectricPowerDemand.time = PeakDemandBound.time
                  AND PeakDemandBound.time = KW.time
            THEN '1' ELSE '0' END) AS Indicator
      FROM PayPeriod, ElectricPowerDemand, PeakDemandBound, KW);
```

STEP 6: Create the parameter learning event and then execute the event construct to learn the parameters (Box 8), which are stored in their tables respectively.

*Box 8.*

```
CREATE EVENT LearnPeakDemandBoundParameter (
      GC_LEARN PeakDemandBound, PayPeriodSupplyDemand, KW
      FOR MINIMIZE SUM(MESC.Charge) AS TotalCharge
      WITH CBM.Indicator = '1' THEN
            CBM.payPeriodSupplyDemand >= CBM.kw
      AND PBM.Indicator = '1' THEN
            PBM.payPeriodSupplyDemand >= 0.9 * PBM.kw
      AND PDB.value <= PPSD.value
      AND PDB.value >= 0
      AND EPGPDB.Indicator = '1' THEN
            EPGPDB.kw = EPGPDB.peakDemandBound
      AND EPLEPDB.Indicator = '1' THEN
            EPLEPDB.kw = EPLEPDB.electricPowerDemand
      FROM CurrentBillingMonth CBM, PrecedingBillingMonth PBM,
```

```
      PeakDemandBound PDB, PayPeriodSupplyDemand PPSD, KW,
      ElectricPowerGreaterPeakDemandBound EPGPDB,
      ElectricPowerLessEqualPeakDemandBound EPLEPDB,
      MonthlyEServiceCharge MESC
      WHERE CBM.time = PBM.time
      AND PBM.time = PDB.time
      AND PDB.time = PPSD.time
      AND PPSD.time = KW.time
      AND KW.time = EPGPDB.time
      AND EPGPDB.time = EPLEPDB.time
      AND EPLEPDB.time = MESC.time);

EXECUTE LearnPeakDemandBoundParameter;
```

This learning query (Box 8) is to learn `PeakDemandBound`, `PayPeriodSupplyDemand`, and `KW` that minimize the `TotalCharge` of the 24 pay periods and satisfy all the six constraints in the `WITH…THEN` clause. For example, the first two constraints denote the *C1* and *C2* of the contractual terms of the GMU electric bill. The last two constraints express the monitoring templates. When the event `LearnPeakDemandBoundParameter` is executed, the command `GC_LEARN` sends the SQL-learning event to the MTSA compiler, where GC stands for "General Class". The compiler transforms this event to the OPL construct, which is then sent to the IBM ILOG CPLEX Optimizer to learn the parameters, e.g., peakDemandBound. The learned parameters are stored in their corresponding tables, e.g., PeakDemandBound. Note that all the parameters are instantiated with the optimal values.

## MTSA QUERY SEMANTICS: PARAMETER ESTIMIATION MODEL

In this section, we formalize the Multivariate Time Series Analytics - Parameter Estimation (MTSA-PE) problem and solution that we propose for the *Parameter Learning Service*. The goal of a MTSA-PE problem is to learn optimal decision parameters that maximize or minimize an objective function over historical and projected multivariate time series data.

We assume that time is split into base time intervals of a fixed duration, e.g., hourly, for simplicity, and each time interval is indexed by an integer, and that we are also given a *m*-sets of decision parameters $\{\overline{w}_1, \overline{w}_2, \ldots, \overline{w}_m\}$. The mathematical components of the MTSA-PE problem and solution and its formulations are shown as follows:

- *Time Horizon*: A time horizon $TH$ is defined as $Z_k = \{t | t \in Z \wedge t \geq k\}$, where $Z$ is a set of integers, $t$ is a time interval in $Z_k$, and $k \in Z$.

More specifically, we use negative or zero integers $t \leq 0$ represented for the past and positive integers $t > 0$ represented for the future time intervals. For example, $Z_{-5}$ is the set of all integers that are greater than or equal to -5. It means that the time horizon starts from the past hourly time interval -5 to the future infinite time interval.

- *Past Time Horizon*: A past time horizon is defined as $PastTH = \{t | t \in Z_k \wedge t \leq 0\}$.
- *Future Time Horizon*: A future time horizon is defined as $FutureTH = \{t | t \in Z_k \wedge t > 0\}$.
- *Period Horizon*: A period horizon $PH$ is defined as $Z_l = \{p | p \in Z \wedge p \geq l\}$, where $Z$ is a set of integers, $p$ is a period in $Z_l$, and $l \in Z$.

We also assume that a sequence of time intervals $t$s in $Z_k$ is grouped into periods, e.g., daily,

weekly, or monthly periods. Each period $p$ contains consecutive time intervals and is also indexed by an integer, where a positive integer $p > 0$ corresponds to the future period, and a negative or zero integer $p \leq 0$ corresponds to the past period.

- *Past Period Horizon*: A past period horizon is defined as $PastPH = \{p | p \in Z_l \wedge p \leq 0\}$.
- *Future Period Horizon*: A future period horizon is defined as $FuturePH = \{p | p \in Z_l \wedge p > 0\}$.

The mapping between a time interval in $TH$ and a period in $PH$ is a function: $riod$: $TH \rightarrow PH$ $such\ that$ $(\forall t_1, t_2 \in TH)$: $(t_1 < t_2 \rightarrow Period(t_1) \leq Period(t_2))$. Now suppose we have a sequence of time intervals in $TH$ and of their corresponding periods in $PH$ that are shown in Figure 3. For instance, both the hourly time intervals, 2 and 3, are mapped to the same period, i.e., $Period(2)$ = $Period(3)$ = 1, as 2 is less than 3 so that 2 and 3 are grouped into the same period 1. Some other examples are $Period(0)$ = 0, $Period(8)$ = 3, and $Period(-6)$ = -2.

*Figure 3. Time Intervals in TH and Corresponding Periods in PH*

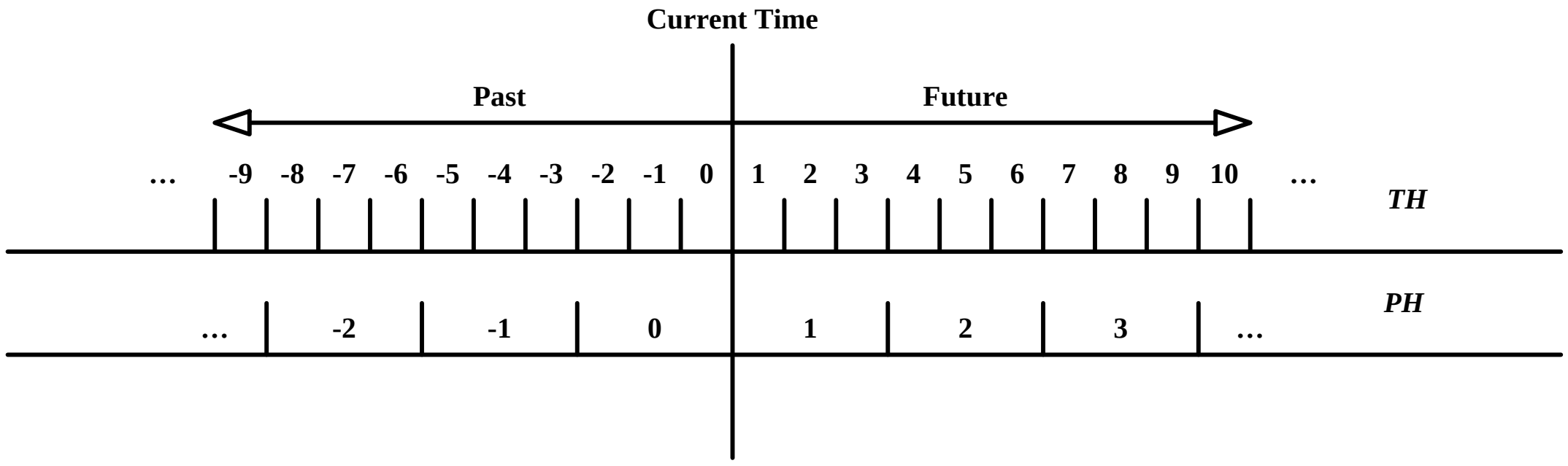

- *Time Series*: A time series $S$ is a function $S$: $TH \rightarrow D$, where $D$ is a numerical domain, e.g., $D = R$ or $D = Z$.
- *Parametric Estimation Constraint*: A parametric estimation constraint $C(S_1(t), S_2(t), \ldots, S_k(t), p_1, p_2, \ldots, p_n)$ is a symbolic expression in terms of $S_1(t), S_2(t), \ldots, S_k(t), p_1, p_2, \ldots, p_n$, where $S_1(t), S_2(t), \ldots, S_k(t)$ are the $k$ input time series, and $(p_1, p_2, \ldots, p_n)$ is a vector of $n$ parameters that have been given by domain experts or are instantiated at a particular time interval over $TH$ or a period over $PH$.

We suppose that a parametric estimation constraint is written in a language that has the truth-value interpretation $I$: $R^k x R^n \rightarrow$ {TRUE, FALSE}, i.e., $I(C(S_1(t), S_2(t), \ldots, S_k(t), p_1, p_2, \ldots, p_n))$ = TRUE if and only if the constraint $C$ is satisfied at $t$ with the parameters $(p_1, p_2, \ldots, p_n) \in R^n$. In this paper, we focus on any possible combinations of inequality constraints of the general form: $(S_1(t)\ eop\ p_1)\ op\ (S_2(t)\ eop\ p_2)\ op\ \ldots\ op\ (S_k(t)\ eop\ p_n)\ op\ (p_1\ eop\ p_2)\ op\ (p_1\ eop\ p_3)\ op\ \ldots\ op\ (p_i\ eop\ p_j)$, where $eop \in \{<, \leq, ==, \geq, >\}$, $op \in \{\wedge, \vee\}$, and $i \neq j$.

- *Parametric Implication Constraint*: A parametric implication constraint ($C_\ell \rightarrow C_j$) is a logical constraint, i.e., "$C_\ell$ implies $C_j$" or "if $C_\ell$ then $C_j$" is TRUE if both $C_\ell$ and $C_j$ are TRUE, where $C_\ell$ and $C_j$ are a parametric estimation constraint $C$.

Some of the $C$s and ($C_\ell \rightarrow C_j$)s are parametric global constraints $\boldsymbol{C_P}$ that are the general constraints, e.g., the contractual terms of the GMU electric bill, such as *C1*, to be applied to an entire problem. Some of them are parametric monitoring constraints $\boldsymbol{C_M}$ that are used to detect the occurrence of an event of a problem, e.g., *electricPowerDemand* > *peakDemandBound*.

- *Objective Function*: An objective function $O$ is a function $O$: $R^{w_1} \mathrm{x} R^{w_2} \mathrm{x} \ldots \mathrm{x} R^{w_m} \rightarrow R$, where $m$ is the total number of sets of parameters, $w_i$ is the total number of parameters in a set *i*, and *R* is the set of real numbers, for *i* = 1, 2, …, *m*.

*MTSA Parametric Estimation* (*MTSA-PE*) *Problem*: A *MTSA-PE* problem is a tuple <***S***, ***P***, $\boldsymbol{C_P}$, $\boldsymbol{C_M}$, ***O***>, where ***S*** = {$S_1$, $S_2$, …, $S_k$} is a set of the *k* input time series, ***P*** = {$\overline{w}_1$, $\overline{w}_2$, …, $\overline{w}_m$} is a *m*-sets of parameters, $\boldsymbol{C_P}$ is a set of parametric global constraints in ***S*** and ***P***, $\boldsymbol{C_M}$ is a set of parametric monitoring constraints in ***S*** and ***P***, and ***O*** is an objective function.

*MTSA Parametric Estimation* (*MTSA-PE*) *Solution*: A solution to the *MTSA-PE* problem <***S***, ***P***, $\boldsymbol{C_P}$, $\boldsymbol{C_M}$, ***O***> is *argmin* ***O***(***P***), i.e., the optimal values of a *m*-sets of parameters that minimize ***O***.

Let us reconsider our GMU example as an explanation for the above mathematical formulations. First, Table 1 shows the input multivariate time series ***S***, and the time interval $t$ is an integer hourly time interval. All the input time series are stored in the tables that we discussed in the STEP 1 of the parameter learning service in the fourth section.

*Table 1. Multivariate Time Series Data S*

| **TIME SERIES *S*** | **ABBREVIATION** | **TABLE** |
|---|---|---|
| Electric Power Demand | $electricPowerDemand(t)$ | ElectricPowerDemand |
| Monthly Pay Period | $payPeriod(t)$ | PayPeriod |
| Annual Year | $year(t)$ | Year |
| Month of a Year | $month(t)$ | Month |
| Day of a Month | $day(t)$ | Day |
| Day of a Week | $weekDay(t)$ | WeekDay |
| Hour of a Day | $hour(t)$ | Hour |

The decision parameter sets ***P*** used in the case study are defined and explained in Table 2, and $p$ is a monthly pay period. All the tables of the parameter sets are created in the STEP 2 of the parameter learning service in the fourth section.

*Table 2. Decision Parameter Sets*

| **PARAMETER SET** | **USAGE INTREPRETATION** | **TABLE** |
|---|---|---|
| $peakDemandBound[p]$ is an array to store the peak demand bound for each monthly pay period. | • Test if the $electricPowerDemand(t)$ exceeds the $peakDemandBound[p]$ when $p == payPeriod(t) \land t \geq 1$ for $\forall t \in TH$ and $p \in PH$.<br>• Test if the $electricPowerDemand(t)$ less than or equal to $peakDemandBound[p]$ when $p == payPeriod(t)$ or $t \leq 0$ for $\forall t \in TH$ and $p \in PH$. | PeakDemandBound |
| $kW[t]$ is an array to store the electric power demand for each hourly time interval. | • Instantiate the values into $kW$ over the historical and projected electric power demand for | KW |
| | $\forall t \in TH, p \in PH$, i.e., $kW[t]$ stores the electric power demand when the electric power demand is less than or equal to the peak demand bound, or $kW[t]$ stores the peak demand bound when the electric power demand is greater than the peak demand bound. | |
| $payPeriodSupplyDemand[p]$ is an array to store the electricity supply demand for each monthly pay period. | • Instantiate the values into $payPeriodSupplyDemand$ over $kW$ depended on which contractual condition *C1* or *C2* in the second section is satisfied such that $payPeriodSupplyDemand$ minimizes the objective function ***O*** for $\forall t \in TH, p \in FuturePH$. | PayPeriodSupplyDemand |

$\boldsymbol{C_P}$ and $\boldsymbol{C_M}$ are illustrated as follows. Both types of the constraints are created in the STEP 4 and STEP 5 of the parameter learning service in the fourth section.

- Parametric global constraint $\boldsymbol{C_P}$

$C_{\ell}$: $C(payPeriod(t), weekDay(t), hour(t), month(t), p, 1,5, 6, 7,9, 10,22) = (payPeriod(t) == p \wedge (weekDay(t) \geq 1 \wedge weekDay(t) \leq 5) \wedge ((hour(t) \geq 10 \wedge hour(t) \leq 22 \wedge month(t) \geq 6 \wedge month(t) \leq 9) \vee ((hour(t) \geq 7 \wedge hour(t) \leq 22) \wedge (month(t) \leq 5 \vee month(t) \geq 10))))$

$C_{j}$: $C(payPeriodSupplyDemand[p], kW[t]) = (payPeriodSupplyDemand[p] \geq kW[t])$

$\boldsymbol{C_1}$: $(\forall t \in TH, p \in FuturePH)$: $C_{\ell} \rightarrow C_{j}$ denotes the contractual condition *C1* in the second section. This condition *C1* is also constructed by the `CBM.Indicator = '1' THEN CBM.payPeriodSupplyDemand >= CBM.kw` that is evaluated by the parameter learning event, i.e., `LearnPeakDemandBoundParameter`, in the STEP 6.

$C_{\ell}$: $C(payPeriod(t), weekDay(t), hour(t), month(t), p - 11, p, 1,5, 6,9, 10,22) = ((payPeriod(t) \geq p - 11 \wedge payPeriod(t) < p) \wedge (weekDay(t) \geq 1 \wedge weekDay(t) \leq 5) \wedge (month(t) \geq 6 \wedge month(t) \leq 9) \wedge (hour(t) \geq 10 \wedge hour(t) \leq 22))$

$C_{j}$: $C(payPeriodSupplyDemand[p], kW[t], 0.9) = (payPeriodSupplyDemand[p] \geq 0.9 * kW[t])$

$\boldsymbol{C_2}$: $(\forall t \in TH, p \in FuturePH)$: $C_{\ell} \rightarrow C_{j}$ represents the contractual condition *C2*, which is constructed by the `PBM.Indicator = '1' THEN PBM.payPeriodSupplyDemand >= 0.9 * PBM.kw`, in the second section.

***C*₃**: $(\forall p \in FuturePH)$: $(peakDemandBound[p] \leq payPeriodSupplyDemand[p])$ restricts the peak demand bound not greater than the electricity supply demand. This constraint is constructed by the `PDB.value <= PPSD.value`.

**$C_4$**: $(\forall p \in FuturePH)$: $(peakDemandBound[p] \geq 0)$ ensures that the peak demand bound must be non-negative values. This constraint is constructed by the `PDB.value >= 0`.

- Parametric Monitoring Constraint $\boldsymbol{C_M}$

  $C_\ell$: $C(payPeriod(t), electricPowerDemand(t), peakDemandBound[p], t, 1, p)$ = $(t \geq 1 \wedge p == payPeriod(t) \wedge electricPowerDemand(t) > peakDemandBound[p])$

  $C_f$: $C(kW[t], peakDemandBound[p])$ = $(kW[t] == peakDemandBound[p])$

**$C_5$**: $(\forall t \in TH, p \in PH)$: $C_\ell \rightarrow C$ monitors whether the electric power demand exceeds the peak demand bound when the hourly time interval $t$ is positive. If this monitoring constraint, $C_\ell$, is triggered, the peak demand bound is stored in the $kW$. This monitoring constraint is constructed by the `EPGPDB.Indicator = '1' THEN EPGPDB.kw = EPGPDB.peakDemandBound` that is evaluated by the parameter learning event, i.e., LearnPeakDemandBoundParameter, in the STEP 6 as well.

  $C_\ell$: $C(payPeriod(t), electricPowerDemand(t), peakDemandBound[p], p, t, 0)$ = $((electricPowerDemand(t) \leq peakDemandBound[p] \wedge p == payPeriod(t)) \vee (t \leq 0))$

  $C_f$: $C(kW[t], peakDemandBound[p])$ = $(kW[t] == electricPowerDemand(t))$

**$C_6$**: $(\forall t \in TH, p \in PH)$: $C_\ell \rightarrow C$ monitors whether the electric power demand is less than or equal to the peak demand bound or the hourly time interval $t$ is non-positive. If this monitoring constraint, $C_\ell$, is triggered, the electric power demand is stored in the $kW$. This constraint is constructed by the `EPLEPDB.Indicator = '1' THEN EPLEPDB.kw = EPLEPDB.electricPowerDemand`

Regarding the objective function ***O***, we assume that the GMU energy planners evaluate the total peak demand charge of the ES service for the projected 24 pay periods in the future two years, i.e., $\sum_{p=1}^{24}(8.124 * payPeriodSupplyDemand[p])$, where $8.124 * payPeriodSupplyDemand[p]$ is the monthly ES service charge, which is created in the STEP 3 of the parameter learning service in the fourth section. This total peak demand charge is the objective function ***O***, which is minimized by optimally determining the $payPeriodSupplyDemand[p]$ that satisfies all the given constraints, where $8.124 is the generation demand charge per kilowatt according to the contractual terms of the GMU electric bill, and $1 \leq p \leq 24$.

Shown in Table 3, the MTSA-PE problem and solution of our example can be constructed by putting all the considered time series ***S***, the parameter sets ***P***, the constraints $\boldsymbol{C_P}$ and $\boldsymbol{C_M}$, and the objective function ***O*** to the formulations of the *MTSA-PE* problem and solution. More specifically, a *MTSA-PE* problem and solution is:

$$\underset{\boldsymbol{P}}{argmin}\ \boldsymbol{O}(\boldsymbol{P})$$
$$subject\ to\ \boldsymbol{C_P}(\boldsymbol{P}) \wedge \boldsymbol{C_M}(\boldsymbol{P})$$

This MTSA-PE problem and solution is constructed by the learning event LearnPeakDemandBoundParameter, which learns the parameter sets, $peakDemandBound$, $payPeriodSupplyDemand$, and $kW$.

*Table 3. Formulation of the MTSA-PE Problem and Solution for the GMU Peak Electric power demand*

| **PROBLEM AND SOLUTION** |
|---|
| **PROBLEM:**<br><***S***, ***P***, ***C<sub></sub>***$\boldsymbol{C_P}$, $\boldsymbol{C_M}$, ***O***><br><br>***S*** = $\{electricPowerDemand, payPeriod, year, month, day, weekDay, hour\}$,<br><br>where $electricPowerDemand(t) \geq 0, -8759 \leq t \leq 17520, -11 \leq payPeriod(t) \leq 24, 2011 \leq year(t) \leq 2013, 1 \leq month(t) \leq 12, 1 \leq day(t) \leq 31, 0 \leq weekDay(t) \leq 6, 0 \leq hour(t) \leq 23$<br><br>***P*** = $\{peakDemandBound, payPeriodSupplyDemand, kW\}$ ,<br><br>where $peakDemandBound[p] \geq 0, payPeriodSupplyDemand \geq 0, kW[t], -8759 \leq t \leq 17520, -11 \leq p \leq 24$<br><br>$\boldsymbol{C_P}$ = $\{C_1, C_2, C_3, C_4\}$, where<br><br>$C_1$ = $(\forall t \in TH, p \in FuturePH): ((payPeriod(t) == p \wedge (weekDay(t) \geq 1 \wedge weekDay(t) \leq 5) \wedge ((hour(t) \geq 10 \wedge hour(t) \leq 22 \wedge month(t) \geq 6 \wedge month(t) \leq 9) \vee ((hour(t) \geq 7 \wedge hour(t) \leq 22) \wedge (month(t) \leq 5 \vee month(t) \geq 10)))) \rightarrow (payPeriodSupplyDemand[p] \geq kW[t]))$,<br><br>$C_2$ = $(\forall t \in TH, p \in FuturePH): (((payPeriod(t) \geq p - 11 \wedge payPeriod(t) < p) \wedge (weekDay(t) \geq 1 \wedge weekDay(t) \leq 5) \wedge (month(t) \geq 6 \wedge month(t) \leq 9) \wedge (hour(t) \geq 10 \wedge hour(t) \leq 22)) \rightarrow (payPeriodSupplyDemand[p] \geq 0.9 * kW[t]))$,<br><br>$C_3$ = $(\forall p \in FuturePH): (peakDemandBound[p] \leq payPeriodSupplyDemand[p])$,<br><br>$C_4$ = $(\forall p \in FuturePH): (peakDemandBound[p] \geq 0)$<br><br>$\boldsymbol{C_M}$ = $\{C_5, C_6\}$, where<br><br>$C_5$ = $(\forall t \in TH, p \in PH): ((t \geq 1 \wedge p == payPeriod(t) \wedge electricPowerDemand(t) > peakDemandBound[p]) \rightarrow (kW[t] == peakDemandBound[p]))$<br><br>$C_6$ = $(\forall t \in TH, p \in PH): (((electricPowerDemand(t) \leq peakDemandBound[p] \wedge p ==$ |

$payPeriod(t)) \vee (t \leq 0)) \rightarrow (kW[t] == electricPowerDemand(t))),$

$\boldsymbol{O} = \sum_{p=1}^{24}(8.124 * payPeriodSupplyDemand[p])$

**SOLUTION:**

$\underset{\boldsymbol{P}}{argmin}\, \boldsymbol{O}(\boldsymbol{P})$

$subject\ to\ \boldsymbol{C_P}(\boldsymbol{P}) \wedge \boldsymbol{C_M}(\boldsymbol{P}))$

This MTSA-PE problem is then expressed by the MTSA SQL according to the STEP 6 of the parameter learning service in the fourth section. Once this MTSA-SQL construct of this problem is initiated, the optimal values of the decision parameter sets $\boldsymbol{P}$ are determined by sending this MTSA-SQL construct to the MTSA compiler. This MTSA compiler transforms the MTSA-SQL construct to the OPL format that is then sent to the external optimization solver, i.e., IBM ILOG CPLEX Optimizer, to learn the parameter sets $\boldsymbol{P}$. After the optimal decision parameters, e.g., $peakDemandBound[1]$ of the monthly pay period 1, is learned from the optimizer, we can apply the parametric monitoring constraints, e.g., $electricPowerDemand(t) > peakDemandBound[1]$, to the new incoming electricity consumption of the monthly pay period 1 from the GMU Fairfax campus and perform the event monitoring in an hourly basis, where $1 \leq t \leq 720$, for the entire monthly pay period 1. Once the monitoring constraints, e.g., $C_5$, are triggered, the recommended action, e.g., the electric load shedding, is alerted to the service providers, that is, the GMU energy planners, to execute the electric load shedding to shut down some electric account units according to the prioritization scheme from the energy manager.

## IMPLEMENTATION OF A HIGH-LEVEL ARCHITECTURE FOR PARAMETER LEARNING PROCESS

Figure 4 illustrates the parameter learning process for the optimal decision parameters. As this figure shows, domain experts use the parameter learning service to construct the query for the MTSA learning event, e.g., `LearnPeakDemandBoundParameter`. Once this learning event is initiated, the MTSA compiler calls the query translator to transform the learning event into the IBM OPL construct, which is shown in Figure 5 of the Appendix section. Note that this OPL construct is manually created as we are working on the compiler. This IBM OPL construct is then sent to the external optimization solver, i.e., the IBM ILOG CPLEX Optimizer, to learn optimal decision parameters, e.g., $peakDemandBound$. These decision parameters are then processed by the output formatter associated with the query translator to return the answer back to the parameter learning service, which presents the results to the experts.

*Figure 4. Parameter Learning Architecture for Optimal Decision Parameters*

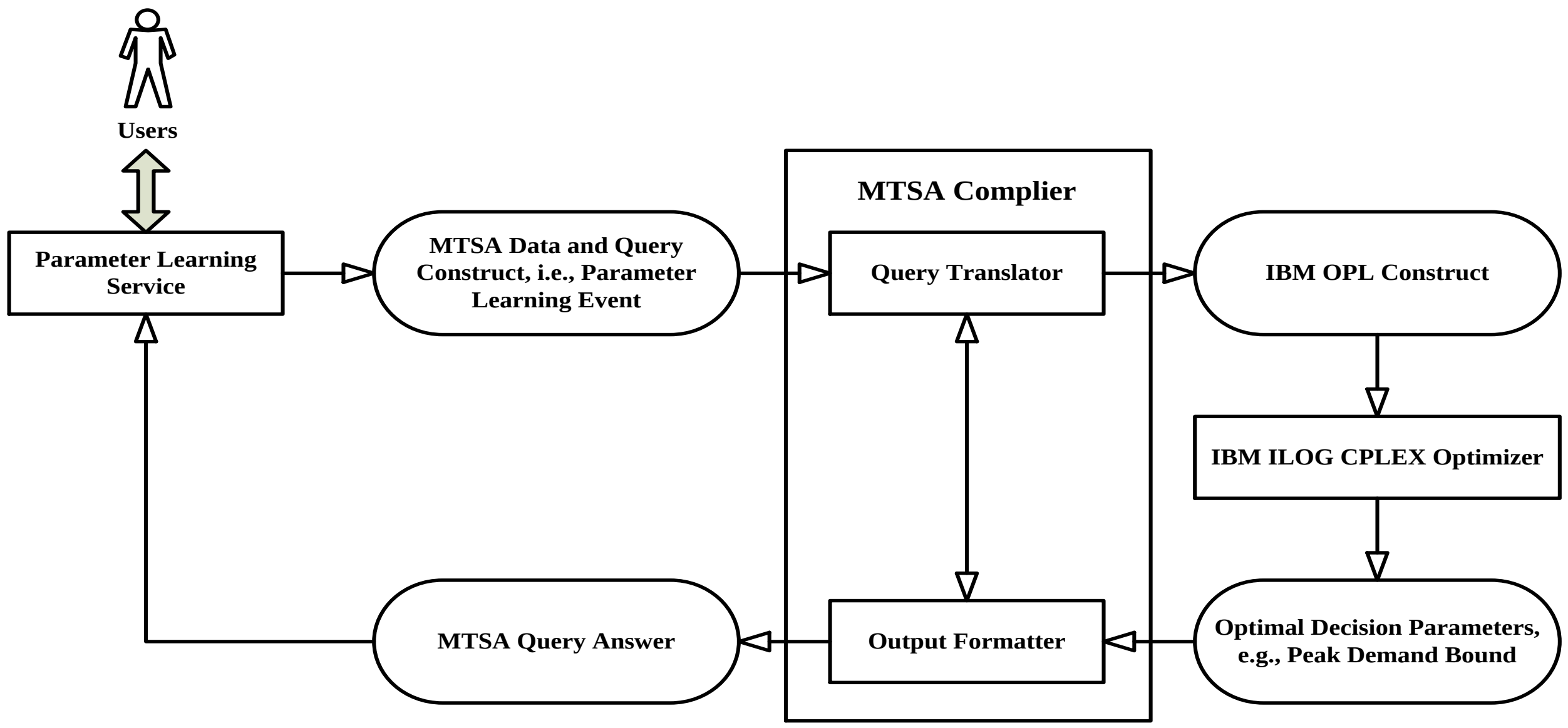

Figure 5 in the Appendix section shows the OPL model transformed from the MTSA-SQL construct. First, the value 24, i.e., the total 24 months from 2012 to 2013, is assigned to the variable nbPayPeriods in the line number 7. The value 0 is assigned to the variable annualBound, that is, the maximal annual power interruption allowed in the line number 8. From the line number 11 to 19, we declare a tuple of a power interval that has the attributes, including pInterval, payPeriod, year, month, day, hour, and weekDay. The line number 21 declares the PowerIntervals that include both the past and the future power intervals. The line number 22 declares the electricPowerDemand[PowerIntervals] array. The line number 24, 25, and 26 declares the parameter sets, including the peakDemandBound[PayPeriods], kW[PowerIntervals], and payPeriodSupplyDemand[PayPeriods]. The monthly ES service charge is declared on the line number 29, and the total peak demand charge, which is declared on the line number 30, is minimized on the line number 32. All the constraints, *C1 – C6*, are declared from the line number 35 to 54.

## EXPERIMENTAL CASE STUDY ON THE PEAK DEMAND BOUND

Using the historical electric power consumption in the past years, e.g., 2011, the projected electricity demand over a future time horizon, e.g., 2012 and 2013, the maximal annual power interruption allowed, e.g., 0, and the parametric model templates, i.e., global and monitoring constraints, which are identified by the GMU energy planners and are required by the utility contracts that are supplied from the *Model Definition Service*, we input all of these data, constraints, and requirements to the *Parameter Learning Service*.

Using the *Parameter Learning Service*, we formulate the MTSA query construct, i.e., the parameter learning event, which has been demonstrated in the fourth section. Based on the parameter learning event, we manually formulate the corresponding OPL construct, shown in the Appendix, and run the construct on the IBM ILOG CPLEX Optimization Studio to obtain the learned optimal peak demand bounds for all the future pay periods shown in Table 4.

*Table 4. Learned Optimal Peak Demand Bounds*

| PAY PERIOD | PEAK DEMAND BOUND kW | PAY PERIOD | PEAK DEMAND BOUND kW |
|---|---|---|---|
| January 2012 | 12189 | January 2013 | 12953 |
| February 2012 | 12654 | February 2013 | 13447 |
| March 2012 | 12268 | March 2013 | 13037 |
| April 2012 | 15410 | April 2013 | 16376 |
| May 2012 | 14729 | May 2013 | 15653 |
| June 2012 | 14921 | June 2013 | 15856 |
| July 2012 | 17211 | July 2013 | 18291 |
| August 2012 | 14575 | August 2013 | 15490 |
| September 2012 | 15998 | September 2013 | 17001 |
| October 2012 | 15020 | October 2013 | 15962 |
| November 2012 | 12856 | November 2013 | 13662 |
| December 2012 | 12654 | December 2013 | 13447 |

The energy planners use the above optimal peak demand bounds to perform the optimal event monitoring over the actual incoming electricity demand in each monthly pay period through the *Monitoring and Recommendation Service* described in the fourth section.

Using the results from the *Monitoring and Recommendation Service*, i.e., when to perform the load shedding, the actual QoS, i.e., the power interruption, the actual cost saving, i.e., the monthly electricity charge, and the corresponding optimal parameters and values as inputs, we evaluate the parametric model templates through the *Model Accuracy and Quality Evaluation Service*. Based upon the differences in terms of the load shedding, the QoS, and the electricity charge, this evaluation module generates a MTSA query construct to update the model templates accordingly. After that, the energy planners can use the updated templates with the input time series, QoS, and requirements to repeat the same process to learn a new set of optimal peak demand bounds for monitoring in the future pay period.

## CONCLUSION AND FUTURE WORK

In this paper, we propose a Web-Mashup Application Service Framework for Multivariate Time Series Analytics (MTSA) that supports the services of model definitions, querying, parameter learning, model evaluations, data monitoring, decision recommendations, and web portals. This framework maintains the advantage of combining the strengths of both the domain-knowledge-based and the formal-learning-based approaches and is designed for a more general class of problems over multivariate time series. More specifically, we identify a general-hybrid-based model, Multivariate Time Series Analytics – Parameter Estimation, to solve this class of problems in which the objective function is maximized or minimized from the optimal decision parameters regardless of particular time points. This model also allows domain experts to include multiple types of constraints, e.g., global constraints and monitoring constraints, as well. We further extend the MTSA data model and query language to support this class of problems for the services of learning, monitoring, and recommendation. At the end, we conduct an experimental case study on the Fairfax campus microgrid at George Mason University to demonstrate our proposed framework, models, and language.

In addition to the contributions made by the research in this paper, there are still numerous interesting questions for further explorations. They include the issues regarding the multi-event MTSA-PE model, the effective algorithms for parameter learning on one- and multi-event MTSA-PE problems, the query language for parameter learning on multi-sequential events, and the development of the Model Accuracy and Quality Evaluation module.

First, the MTSA-PE model is designed to solve a class of problems that involve a single event, e.g., the electric load shedding is executed when the electricity demand exceeds the optimal peak demand bound. However, there are many real-world cases that multiple related events occur in sequence. For instance, consider the above load-shedding example again, in which the energy planners would like to determine when the electric account units should be turned off and when those accounts should be turned on in order. Using the MTSA-PE model, the energy planners would not be able to properly decide on when to shed and unshed the load at the two interrelated events and then to gain the maximal cost savings and achieve the minimal power interruptions. To address the shortcomings of this issue, the future research will focus on developing a multi-event MTSA-PE model. This new model will maintain the advantages of the single-event MTSA-PE model and also support the parameter learning on multiple events in sequence.

Second, the IBM ILOG CPLEX optimizer that we use to solve the class of MTSA-PE problems is the branch-and-bound-based algorithm, which the time complexity is exponential, i.e., $O(k2^N)$ if the problem is a single event, and $O(k2^{m \cdot N})$ if the problem is a sequence of multiple events. Thus the future research will focus on developing a new algorithm that will be able to solve the class of single- and multi-event MTSA-PE problems at a lower computational cost.

Third, the proposed MTSA query language is only able to formulate a single-event parameter learning on the MTSA-PE problems. The future research will be how to extend the proposed MTSA query language to support the formulation of the multi-event MTSA-PE problems.

Finally, in order to improve the accuracy of the parametric model templates and the quality of the monitoring and recommendation service in the future, we will develop an evaluation model and algorithm to identify the performance gaps in terms of QoS, event utility, decision-making actions, etc. We will also develop a MTSA query construct to update the model template based on those performance gaps.

# APPENDIX

*Figure 5. The OPL Constructs for the MTSA-SQL Parameter Learning Service*

Peak Demand Model.mod

```
/*********************************************
 * OPL 12.4 Peak Demand Model                *
 * Author: Alexander Brodsky and Chun-Kit Ngan*
 * Creation Date: Dec 11, 2012 at 8:28:56 PM  *
 *********************************************/
float timeIntervalSize = ...;
int nbPayPeriods = ...;
float annualBound = ...;
range PayPeriods = 1..nbPayPeriods;

tuple powerInterval{
  int pInterval;
  int payPeriod;
  int year;
  int month;
  int day;
  int hour;
  int weekDay;
}

{powerInterval} PowerIntervals = ...;
float electricPowerDemand[PowerIntervals] = ...;

dvar float+ peakDemandBound[PayPeriods];
dvar float+ kW[PowerIntervals];
dvar float+ payPeriodSupplyDemand[PayPeriods];

pwlFunction kWfunction[i in PowerIntervals] = piecewise{1 -> electricPowerDemand[i]; 0};
dexpr float generationDemandCharge[p in PayPeriods] = 8.124 * payPeriodSupplyDemand[p];
dexpr float totalCharge = sum(p in PayPeriods) (generationDemandCharge[p]);

minimize totalCharge;

subject to {
   forall(i in PowerIntervals : i.pInterval <= 0) kW[i] == electricPowerDemand[i];

   forall(i in PowerIntervals : i.pInterval >= 1) kW[i] == kWfunction[i](peakDemandBound[i.payPeriod]);

   forall (p in PayPeriods) peakDemandBound[p] <= payPeriodSupplyDemand[p];

   forall(p in PayPeriods)
       forall(i in PowerIntervals : i.payPeriod == p && i.weekDay >= 1 && i.weekDay <= 5
               && ((i.month >= 6 && i.month <= 9 && i.hour >= 10 && i.hour <= 22) ||
               (i.month <= 5 && i.month >= 10 && i.hour >= 7 && i.hour <= 22)))
                   payPeriodSupplyDemand[p] >= kW[i];

   forall(p in PayPeriods)
       forall(i in PowerIntervals : i.month >= 6 && i.month <= 9 && i.payPeriod >= p - 11
               && i.payPeriod < p && i.payPeriod < p && i.weekDay >= 1 && i.weekDay <= 5
               && i.hour >= 10 && i.hour <= 22)
                   payPeriodSupplyDemand[p] >= 0.9 * kW[i];

   sum(i in PowerIntervals : i.pInterval >= 1) (electricPowerDemand[i] - kW[i]) <= annualBound * 2;
}
```